\documentclass[10pt,twocolumn,letterpaper]{article}

\usepackage{cvpr}              % To produce the CAMERA-READY version
\usepackage{multirow}
\newcommand{\difwy}[1]{{\color{black}#1}}
\newcommand{\M}{AOMGen}

\definecolor{cvprblue}{rgb}{0.21,0.49,0.74}
\usepackage[pagebackref,breaklinks,colorlinks,allcolors=cvprblue]{hyperref}

\def\paperID{****} % *** Enter the Paper ID here
\def\confName{CVPR}
\def\confYear{2026}

\title{AOMGen: Photoreal, Physics-Consistent Demonstration Generation for Articulated Object Manipulation}

\author{Yulu Wu
\and
Jiujun Cheng
\and
Haowen Wang
\and
Dengyang Suo
\and
Pei Ren
\and
Qichao Mao
\and
Shangce Gao
\and
Yakun Huang
}

\begin{document}
\twocolumn[{
\maketitle
\vspace*{-0.15in}
\centering
\includegraphics[width=0.9\linewidth, page=1,clip, trim=0 0 0 0]{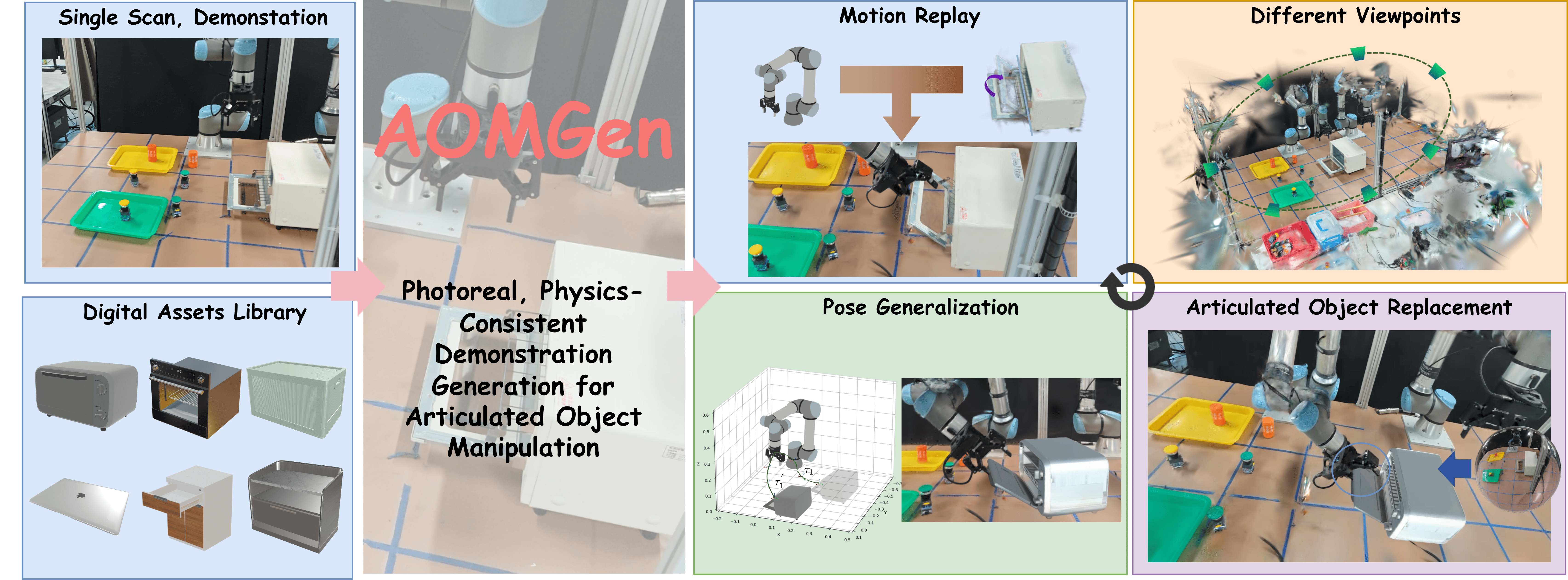}
\captionof{figure}{As a powerful articulated object manipulation data generator, the proposed AOGen generates visually realistic and interaction-accurate data for any object of the same category within a unified framework. At the same time, the generated data provides effective assistance in improving the model's performance.
    }
\vspace{0.25in}
\label{fig:fig1}
}
]

\begin{abstract}
 % Manipulation data obtained from teleoperated demonstrations face challenges such as high costs and limited data diversity. 
 % However, when dealing with articulated objects, existing methods are limited by inaccurate physical simulations, insufficient visual realism, or lack of generalization capability. 
 % This paper proposes AOMGen, a novel method that generates diverse, visually realistic articulated object manipulation data, including rotational joint and prismatic joint, by directly manipulating 3D Gaussians. 
 % In detail, through supervision of the robotic arm's trajectory, AOMGen enables the accuracy motion replay of the original object and the replacement of any articulated object of the same category, thus achieving data generalization. 
 % Moreover, visual augmentation methods, including Gaussian Inpainting and Lighting Application, are applied to ensure that the rendered demonstration data is realistic. 
 % Comprehensive experiments demonstrate that AOMGen significantly improves the success rate of VLA models in handling articulated objects.
 % Notably, the VLA model fine-tuned with data generated by AOMGen shows a task success rate increase from 0 to 85\%. Moreover, thanks to the powerful data generalization capability of AOMGen, the robustness of the VLA model is significantly improved.
Recent advances in Vision-Language-Action (VLA) and world-model methods have improved generalization in tasks such as robotic manipulation and object interaction. However, Successful execution of such tasks depends on large, costly collections of real demonstrations, especially for fine-grained manipulation of articulated objects. To address this, we present AOMGen, a scalable data generation framework for articulated manipulation which is instantiated from a single real scan, demonstration and a library of readily available digital assets, yielding photoreal training data with verified physical states. The framework synthesizes synchronized multi-view RGB temporally aligned with action commands and state annotations for joints and contacts, and systematically varies camera viewpoints, object styles, and object poses to expand a single execution into a diverse corpus. Experimental results demonstrate that fine-tuning VLA policies on AOMGen data increases the success rate from 0\% to 88.7\%, and the policies are tested on unseen objects and layouts. 
 \end{abstract}  
% \begin{figure*}[t]  
%     \centering  
%     \includegraphics[width=1\textwidth]{Fig/figure2.png} 
%     \caption{As a powerful articulated object manipulation data generator, the proposed AOGen generates visually realistic and interaction-accurate data for any object of the same category within a unified framework. At the same time, the generated data provides effective assistance in improving the model's performance.} 
%     \label{fig:fig1} 
% \end{figure*}
\section{Introduction}
\label{sec:intro}

% generalist robot policies is hot -> data is important. And articulated object manipulation is important in generalist policies. 
%两个点：1. 铰链物体的相关研究介绍 2. 基于GS的数据生成方案。
% The collection of high-quality manipulation data has emerged as a key prerequisite for building generalist robot policies capable of manipulating diverse objects~\cite{intelligence2025pi05visionlanguageactionmodelopenworld, gr00tn1_2025, kim24openvla}. However, the limitations of efficiency and diversity in human-collected data constrain the development of foundation policy models. 
% Robotic manipulation models with visuomotor policy are still primarily driven by vast vision-action data, significantly enhancing generalization capabilities~\cite{intelligence2025pi05visionlanguageactionmodelopenworld, gr00tn1_2025, kim24openvla}.
% Robotic manipulation models powered by Vision-Language-Action~(VLA) data have demonstrated remarkable generalization and decision-making capabilities across various tasks.~\cite{intelligence2025pi05visionlanguageactionmodelopenworld, gr00tn1_2025, kim24openvla}.
Robotic manipulation models based on Vision-Language-Action~(VLA), which integrate visual perception, language understanding, and action execution, have demonstrated remarkable generalization capabilities across various tasks~\cite{intelligence2025pi05visionlanguageactionmodelopenworld, gr00tn1_2025, kim24openvla}.
% For fine-grained manipulation, such as handling articulated objects, the demand for demonstration data is even greater. The high cost of human collection makes automated data generation an urgent need. 
% However, such performance depends on large-scale, high-quality human demonstration data.
Achieving such capabilities typically requires substantial amounts of high-quality robot manipulation demonstration data collected from real-world interactions~\cite{open_x_embodiment_rt_x_2023, wu2025robomind, jones24fuse}. 
% For fine-grained manipulation, such as manipulation of constraint-complex articulated objects, more numerous and higher-quality demonstration data are required to learn effective policies. 
In particular, fine-grained manipulation tasks involving articulated objects rely more heavily on diverse and precise data compared to other tasks, 
as their complex kinematic constraints require richer demonstrations for learning effective control policies~\cite{kerr2024rsrd, xue2025demogen}.
% Due to the high cost, low efficiency, and limited coverage of human data collection and annotation, automated, high-fidelity data generation has become an urgent problem that needs to be addressed.
Nevertheless, collecting sufficient high-quality demonstration data remains expensive, labor-intensive, and inherently limited in scenario coverage, making it challenging to obtain adequate data for these tasks.
% To address the scarcity and slow collection of demonstration data, researchers have proposed two main approaches to generate demonstration data. 

To overcome the limitations posed by the inefficient collection of real-world robot manipulation data, 
existing research has primarily explored physics-based simulation and video-driven world modeling as two mainstream strategies.
% On one hand, physics engine-based simulators can batch generate demonstrations with aligned states and actions, significantly reducing the cost of human annotation.~\cite{makoviychuk2021isaacgymhighperformance, Genesis, Wang_2025_CVPR}.
% \difhw{
% One advantage of physics engine-based simulation platforms~\cite{1,2,3} is their ability to generate large batches of demonstrations with aligned states and actions, 
% thereby significantly reducing both the need for manual annotation and the costs associated with robot hardware and physical trial execution.~\cite{1,2,3,4}
% }
% However, simulators are limited by the quality of simulation assets. 
% Although Real-to-Sim-to-Real pipelines~\cite{li2024robogsimreal2sim2realroboticgaussian, jia2025discoverseefficientrobotsimulation} can partially reduce the visual discrepancy between simulated and real environments, 
% they still fail to capture realistic physical interactions, which is crucial for fine-grained manipulation. 
% With the help of Real-to-Sim pipelines~\cite{li2024robogsimreal2sim2realroboticgaussian, jia2025discoverseefficientrobotsimulation}, the visual discrepancy between simulated and real environments can be partially reduced. However, simulators fail to perfectly replicate the complex lighting, materials, and perspectives of the real world.
% \difhw{
% Some methods have achieved promising results by directly training models within simulation environments or utilizing data generated by these simulation platforms~\cite{1,2,4,5}.
% }
Physics-based simulation platforms~\cite{makoviychuk2021isaacgymhighperformance, Genesis, Wang_2025_CVPR} efficiently generate large, state–action aligned demonstrations, reducing manual annotation and hardware costs~\cite{li24simpler, long2025surveylearningembodiedintelligence}, which have enabled several methods to achieve promising results by leveraging simulated data.
However, simulation environments still exhibit significant gaps in visual realism compared to real-world scenes, posing substantial challenges for Sim-to-Real transfer~\cite{li2024robogsimreal2sim2realroboticgaussian, jia2025discoverseefficientrobotsimulation}.
Recent video world model approaches enhance data realism by directly learning from large-scale real-world videos.~\cite{jang2025dreamgen, zhu2025uwm, zhou2024robodreamerlearningcompositionalworld, yang2025orv} Compared with purely simulation-based methods, they offer higher visual fidelity and diversity while reducing reliance on handcrafted simulation assets. However, these models often provide insufficient supervision over physical realism and action executability, leading to physically inconsistent interactions. Hence, for generating visually realistic and physically stable demonstration data, DemoGen~\cite{xue2025demogen} and R2RGen~\cite{xu2025r2rgenrealtoreal3ddata} synthesize additional robot manipulation demonstrations from limited real-world examples, including spatially augmented end-effector trajectories and 3D visual observations. 
However, they have several critical limitations:
(1) limited to simple grasping and placement, unable to handle fine-grained articulated manipulation;
(2) fixed object appearances and geometries, restricting generalization to novel objects or poses;
and (3) single-view inputs, reducing visual realism in multi-view observations.
These limitations underscore the need for methods capable of generating more visually realistic and physically consistent demonstrations for articulated object manipulation tasks.

To address the above limitations, we propose~\M, a novel framework that synthesizes photorealistic and physically consistent demonstration data for category-level articulated object manipulation from a single real-world scan and demonstration (See Figure~\ref{fig:fig1}).
Unlike prior methods,~\M~generalizes across spatial variations and enables object replacement within the same category, enabling significant scalability.
By exploiting the shared kinematic and motion structures among category instances,~\M~transfers manipulation behaviors to novel objects while maintaining visual realism and physical plausibility, without relying on physics simulators.
%对应下文的方法，方法的提出解决了什么问题
% Under the 3DGS representation, achieving realistic physical interaction without relying on a simulator is challenging. AOMGen consists of two modules: 1. High-fidelity reconstruction and motion recovery of real-world manipulation scenes. To achieve physical realism, first, we segment the Gaussian points and align the GS coordinate system with the world coordinate system; then, we fully leverage the information from real-world manipulation data as prior knowledge to achieve motion recovery of the original articulated object. This module achieves geometric-consistent scene reconstruction and motion recovery. 2. Replacement of new objects within the same category and simulating their motion. To adapt to the real world, the replacement articulated object forms a mapping with the real object, and is modeled, including its size, initial state, and motion process, to ensure accurate interaction. At the same time, to further enhance the visual realism of the replaced articulated object, we extract lighting information from the real scene to enhance the material appearance of the replacement object.
~\M~consists of two core modules: 
(1)~\textit{Scene Reconstruction and Motion Recovery}, which accurately reconstructs real-world manipulation scenes using 3D Gaussian Splatting~(3DGS). 
In this module, we first segment Gaussian points from raw observations and align the 3DGS reconstruction with the real-world coordinate frame.
Using real manipulation trajectories as physical priors, we then recover accurate and physically consistent articulated motions, ensuring high-fidelity geometric alignment.
(2)~\textit{Articulated Object Replacement with Pose Generalization}, which enables replacing original objects with new instances from the same category and simulating their corresponding interactions. 
Specifically, we establish mappings between the original and new object models to ensure correct articulation parameters, including joint configurations, sizes, and initial poses. 
We further enhance realism by transferring scene lighting and materials from the original scene to the new objects, producing visually realistic and physically plausible demonstrations for training manipulation policies.
In summary, our main contributions include:
\begin{itemize}
    % \item AOMGen can generate manipulation data for any articulated object of the same category from a static scan video.
    \item Using a single static scan video of an articulated object, AOMGen can generate manipulation data for any other object in the same category.
    % \item AOMGen makes efforts in ensuring physical interaction accuracy and visual realism.
    \item AOMGen ensures precise physical interactions and high visual realism in all the synthesized data.
    % \item The pose of the target object in AOMGen can be arbitrarily generalized, further expanding the generalization boundary of data generation. 
    \item The architecture supports arbitrary adjustments to the target object’s pose, greatly expanding the diversity of configurations in the generated data and pushing the boundaries of generalization.
    % \item The data generated by AOMGen is proven effective in VLA training and can help improve model performance.
    \item The synthetic data produced by AOMGen has proven effective for VLA training, resulting in improved model performance.
\end{itemize}
\section{Related Work}
\label{sec:related_work}

\subsection{Robot Manipulation Data Generation}
The efficient generation of high-quality robotic manipulation data is widely studied by researchers, due to the importance of visual data for policy learning.

MimicGen and its extensions~\cite{mandlekar2023mimicgendatagenerationscalable, hoque2024intervengeninterventionaldatageneration, jiang2025dexmimicgenautomateddatageneration} reconfigure the collected demonstration data to synthesize corresponding manipulation data. This provides a good generalization solution for seen objects; however, it fails to transfer to manipulation cognition for unseen objects.

Another line of work leverages LLM and VLM to generate manipulation data from a single image~\cite{chen2023genaug, jang2025dreamgen, zhu2025uwm, zhou2024robodreamerlearningcompositionalworld, yang2025orv} or generalize existing observation data~\cite{zhou2024autonomous}. These methods have been effective in visual authenticity and efficiency, but due to the uncontrollability of the generative models, the physical authenticity of the generated data cannot be guaranteed. Therefore, AOMGen reconstructs and generalizes one real-world collected data to ensure both its visual and physical performance are excellent.

\subsection{Gaussian Splatting Reconstruction \& Editing}
3D Gaussian Splatting, as an explicit radiance field representation, possesses real-time rendering and interpretable editing capabilities. Some Real-to-Sim-Real pipelines leverage 3DGS to reconstruct scenes and reduce visual discrepancy with the real world~\cite{wu2025rlgsbridge3dgaussiansplatting, lou2024robogsphysicsconsistentspatialtemporal, han2025re3simgeneratinghighfidelitysimulation, torne2024reconcilingrealitysimulationrealtosimtoreal}. Furthermore, to achieve physically realistic interactions, some works directly edit Gaussian points to reconstruct manipulation trajectories consistent with real-world demonstrations from humans~\cite{kerr2024rsrd, yu2025real2render2realscalingrobotdata} or robots~\cite{qureshi2024splatsimzeroshotsim2realtransfer, robosplat}. Specially, Sizhe Yang et al.~\cite{robosplat} propose RobotSplat that leverages 3DGS to generate novel demonstrations. But RobotSplat fails to generalize manipulation data for articulated objects. 

Unlike the robot arm trajectory, the motion trajectory of the target object has no real recorded data for direct reference. Justin Kerr et al.~\cite{kerr2024rsrd} provide a part-level motion recovery method based on DINO features, but how to accurately transfer it to an unseen object remains an open problem. Therefore, we propose a motion transfer method that leverages the real robot arm trajectories to supervise Gaussian editing, thereby generating physically realistic and generalized robotic manipulation data.

On the other hand, Gaussian Editing~\cite{chen2023gaussianeditor} and Inpainting~\cite{huang20253d} is the key factor that determines the visual quality of the generated data. How to correctly segment the Gaussian points of the object to be replaced and enhance the material appearance of the new object to make it more consistent with the real-world environment will determine the rendering realism after Gaussian editing.

\subsection{Articulated Objects Model}
\label{related::ao}
Compared to rigid objects, the motion of articulated objects is constrained by the joints, making their motion patterns more complex than simple transformations in the SO(3) space. Recent works~\cite{10610171, liu2025videoartgsbuildingdigitaltwins} model articulated objects by using multi-view RGB images captured before and after interaction, thereby obtaining the geometric structure of articulated objects (such as the pose of the joints and different parts). However, the above methods fail in the absence of enough visual observations after interaction. Therefore, we propose a training-free articulated object modeling approach that explicitly captures a series of features by analyzing its geometric structure and incorporating robotic arm manipulation data.
\section{\difwy{Problem Definition}}
Given real articulated object manipulation data, which includes static scene scans $V_{static}$, dynamic manipulation video $V_{dynamic}$ and robotic arm joint states $A=\{A_i | i=1,2,\cdots,T\}$, where $T$ represents the total number of time frames in the process, along with several simulation 3D assets of the same category, the goal is to reconstruct the motion of the real manipulation scene and quickly replace the target object in the scene, as well as generalize its pose in any way to generate new observation-action pair manipulation data. We define articulated objects of the same category as objects that have similar relative joint positions within the overall object (e.g., all joints are positioned below of the object) and exhibit the same joint motion patterns. 
\section{Method}
\begin{figure*}[t]  
    \centering  
    \includegraphics[width=1\textwidth]{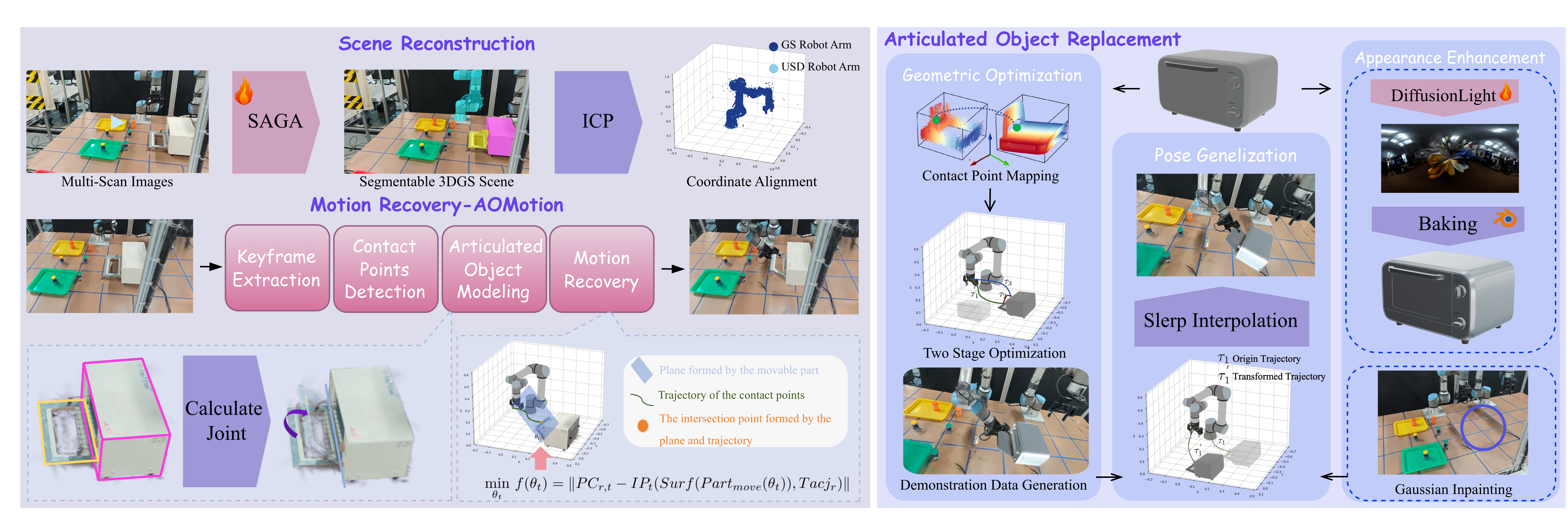} 
    \caption{Pipeline of the proposed AOMGen, where a rotational joint object is used as an example to illustrate the complete pipeline, while a prismatic object can be handled in the same manner.}
    \label{fig:pipeline} 
\end{figure*}

AOMGen can generate abundant articulated object manipulation data with the help of a single demonstration manipulation data. An overview of our
pipeline is shown in Fig~\ref{fig:pipeline}. In this section, we describe AOMGen in detail. First, we prepare for the reconstruction and pre-processing of the static scene based on 3DGS in Sec~\ref{subsec:static_scene_rec}. After preparing the static scene for Gaussian reconstruction, we segment and model the following in Sec~\ref{subsec:motion_replay}: 1) the target articulated object; 2) the robot arm. Using the known trajectory of the robotic arm, we obtain the motion patterns of the articulated object, thereby generating the dynamic scene represented by Gaussian points. Finally, in Sec~\ref{ao_replace}, we build a mapping between the original articulated object and the replacement object, and perform a series of optimizations at the visual level, to generate the generalized demonstration data.

% Scene Reconstruction and Motion Recovery
\subsection{Scene Reconstruction}
\label{subsec:static_scene_rec}
To obtain a high-fidelity reconstruction of the scene, we capture a variety of perspectives of the manipulation scene. Once the images are ready, we use COLMAP to obtain sparse scene reconstruction and camera pose estimation.
% To edit the GS scene, we need to implement part-level GS point segmentation of the target articulated object during the 3DGS training. Based on SAGA~\cite{cen2023saga}, each Gaussian point is attached with a feature $f^D$ of dimensions $D$, which is learned from multiview
% 2D masks extracted by SAM2~\cite{ravi2024sam2}. By adjusting the segmentation granularity of SAM2, part-level image masks can be obtained.

% After obtaining the segmentable 3DGS model $G$, the current 3DGS is in the GS coordinate system, and we need to align it with the real-world coordinate frame to enable simulation of robotic arm motion and scene editing. In detail, given the robot URDF in the real-world coordinate frame, we can sample point clouds $P_{urdf}$ from the mesh surface. Using Iterative Closest Point (ICP)~\cite{zhang2021fast} on $P_{urdf}$ and $P_{robot}$, which are the robot points segmented from $G$, we can get the transformation matrix $T_{gs\rightarrow real}$ and the transformed 3DGS scene $G^{'}$. 

% Based on the existing segmentable GS model, we can obtain the different parts of the articulated object $Part_i \in G$. According to the position of the end effector at different times, distinguish between the movable part $Part_{move}$ and static parts $Part_{static}$.

\subsubsection{Part-level Segmented 3DGS}
Fine-grained segmentation is a necessary condition for editable GS scenes. Based on SAGA~\cite{cen2023saga}, each Gaussian point is attached with a feature $f^D$ of dimensions $D$, which is learned from multi-view 2D masks extracted by SAM2~\cite{ravi2024sam2}. By adjusting the segmentation granularity of SAM2, part-level image masks can be obtained.

Based on the segmented GS model $G$, we can obtain the different parts of the articulated object $Part_i \in G$. According to the position of the end effector at different times, distinguish the movable part $Part_{move}$ and static parts $Part_{static}$ from $Part_i$.

\subsubsection{Coordinate Alignment}
The current GS coordinate system should be aligned with the real-world coordinate frame to ensure consistency in real-world, simulation and GS scenes. Given the robot URDF in the real-world coordinate frame, we can sample point clouds $P_{urdf}$ from the mesh surface. Using Iterative Closest Point (ICP)~\cite{zhang2021fast} on $P_{urdf}$ and $P_{robot}$, which are the robot points segmented from $G$, we can get the transformation matrix $T_{gs\rightarrow real}$ and the transformed 3DGS scene $G^{'}$. 
% Motion Recovery
\subsection{Motion Recovery}
\label{subsec:motion_replay}
% To achieve motion replay, it is necessary to simulate two key objects: the robotic arm and the articulated object. The recovery of the robotic arm is straightforward. Since real-world operation data includes the joint state of the robotic arm, we can segment $P_{robot}^{'} = T_{gs\rightarrow real}P_{robot}$ using the bounding boxes of different links in the robot URDF, into $P_{robot}^{l'}$, where $l$ represents the $l$th of the robot links. Let the initial and joint states at time $t$ be $A_{0}$ and $A_{t} \in A$, respectively. Using Forward Kinematics, the rolation transformation matrix $T_{t}^{l}(A_{0}, A_{t})$ for link $l$ can be obtained. The Gaussian point of the l-th link of the robotic arm at time t is denoted as $P_{robot, t}^{l'}=T_{t}^{l}*P_{robot}^{l'}$
Motion recovery involves simulating two aspects: the robotic arm and the articulated object.

To recover the motion of the robotic arm, the GS robotic arm are segmented using the bounding boxes of different links in the robot URDF, defined as $P_{robot}^{l'}=T_{gs\rightarrow real}P^{l}_{robot}$, where $l$ represents the $l$th link of the robot arm. Given the collected demonstration, including the joint states, the rotation transformation matrix $T_{t}^{l}(A_{0}, A_{t})$ for link $l$ can be calculated by Forward Kinematics, where $A_{0}$ and $A_{t}$ represent the initial and joint states at time $t$.

% On the other hand, due to the lack of real records for the transformation parameters of the articulated object's movable part, we design a supervised method \textbf{AOMotion}, based on the robot arm's trajectory, to recover the motion of the articulated object in the GS scene. AOMotion consists of four submodules: 1. motion key timestamp analysis; 2. contact point analysis; 3. articulated object modeling; 4. motion replay of the movable part. These modules will be described in the following.
On the other hand, to recover the articulated object's motion without value recording, we design a supervised method \textbf{AOMotion}, based on the real-world interaction. AOMotion consists of four submodules: 1. keyframe extraction; 2. contact point detection; 3. articulated object modeling; 4. movable part motion recovery. These modules are described in the following.

\subsubsection{Keyframe Extraction}
% To analyze motion and interaction more precisely, we need to extract the start and end times of the interaction between $P_{robot}^{'}$ and $Part_{move}$ throughout the entire process. Leveraging the dynamic manipulation video $V_{dynamic}$, use SAM2 to generate the mask of the movable part $M_{dynamic}^{t}$ and the robot arm $M_{robot}^{t}$. We aim to determine whether the movable part has started or stopped moving by comparing the difference between $M_{dynamic}^{t}$ and $M_{dynamic}^{t+1}$. However, since $M_{robot}^{t}$ affects the shape of $M_{dynamic}^{t}$, the mask needs to be processed as follows during the comparison:
Leveraging the dynamic manipulation video $V_{dynamic}$, we use SAM2 to generate the mask of the movable part $M_{dynamic}^{t}$ and the robot arm $M_{robot}^{t}$ at frame $t$. We aim to determine whether the movable part starts or stops interacting by comparing the change between $M_{dynamic}^{t}$ and $M_{dynamic}^{t+1}$. However, since $M_{robot}^{t}$ affects the shape of $M_{dynamic}^{t}$, the mask needs to be processed as follows during the comparison:
\begin{equation}
\begin{aligned}
    &M^t &= M_{dynamic}^{t} - M_{robot}^{t+1} \\
    &M^{t+1} &= M_{dynamic}^{t+1} - M_{robot}^{t},
\end{aligned}
\end{equation}
where $M^{t}$ represents the processed mask. Then, we define the following \textbf{Motion Score} to measure whether the current frame has started or finished moving:
\begin{equation}
\begin{aligned}
&d_t(p) =
\begin{cases}
0, & \text{if $p$ in $M^t\&M^{t+1}$ } \\
1, & \text{otherwise}
\end{cases} \\
&MotionScore_t = \frac{\sum_{p \in M^t} d_t(p)}{\sum_{p \in M^t} 1},
\end{aligned}
\end{equation}
where $p$ represents the pixel in the mask, $d_t(p)$ represents whether the pixels within the mask are changing.

Moreover, due to the slight discrepancies in the SAM2 segmentation across different frames, we apply the Savitzky-Golay filter to smooth the motion scores and reduce the noise, resulting in the smoothed motion scores $smooth\_score_t$. Dynamic threshold is used to distinguish between motion and stillness. The baseline $B$ for the motion scores is calculated from the 20th quantile of the motion scores. The noise standard deviation $\sigma_{noise}$ is calculated based on the standard deviation of the motion scores. The final threshold is $\mu = B + 3\sigma_{noise}$.

Based on the smoothed motion scores and dynamic thresholds, we generate a motion label for each frame, and get the first and the last motion frame: 
\begin{equation}
\label{timestamp}
\begin{aligned}
&\text{motion\_frames}_t = 
\begin{cases}
1, & \text{if } smooth\_score_t > \mu \\
0, & \text{otherwise}
\end{cases} \\
&\text{start\_frame} = \min\{t \mid \text{motion\_frames}_t = 1\} \\
& \text{end\_frame} = \max\{t \mid \text{motion\_frames}_t = 1\}
\end{aligned}
\end{equation}

\begin{figure}[b]  
    \centering  
    \includegraphics[width=0.45\textwidth]{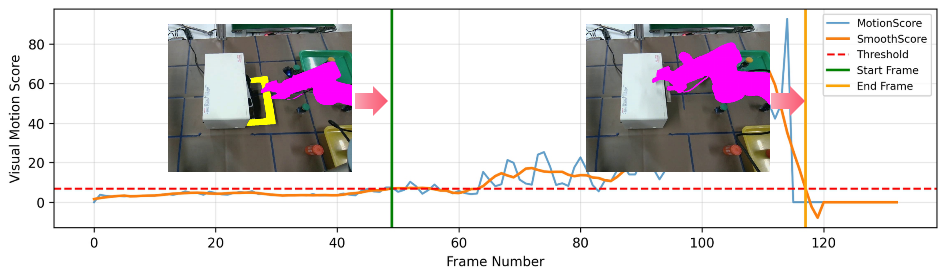} 
    \caption{Computation of the Motion Score.} 
    \label{fig:motionscore} 
\end{figure}

Figure~\ref{fig:motionscore} presents the computation results of the Motion Score.

\subsubsection{Contact Point Detection}
\label{contact_point}
Obtaining the contact point at the beginning of the interaction between the robotic arm and the movable part will help us translate the motion of the articulated object into a rotational transformation at the contact point.

% Given $start\_frame$ calculated in Eq~\ref{timestamp}, we can get the pose of the end-effector at $start\_frame$, defined as $\mathbf{v_{ee}}$. We can obtain the contact point information on the end-effector and the movable part as follows:
The pose of the end-effector at $start\_frame$, which is given by Eq~\ref{timestamp}, is defined as $\mathbf{v_{ee}}$. We can obtain the contact point information on the end-effector and the movable part as follows:
\begin{equation}
PC_{r}= \mathop{\arg\min}\limits_{PC_{r} \in P_{r, t}^{l'}} (\min_{PC_{m} \in Part_{move}}||PC_{robot}-PC_{move}||),
\end{equation}
where $PC_{r}$ represents the contact point on the robot arm. The contact point on the moveable part $PC_{move}$ can be calculated as the same method.

\subsubsection{Articulated Object Modeling}
Given $Part_{move}$ and $Part_{static}$, obtaining the accurate joint direction and center is key. First, construct the bounding boxes of the two parts, $B_{move}=\{e_{move,i} | i\in[1,8]\}$ and $B_{static}=\{e_{static,i} | i\in[1,8]\}$, where $e_i$ stands for the edge of the box. 

In fact, the center and direction of the joint are typically closely related to the geometric position and orientation of the edges to which it is connected. Through looping through all pairs ${e_{static,i}, e_{move,j}|i,j=1,2,\cdots,8}$, we design the following scoring scheme to determine the paired edges adjacent to the joint:
\begin{equation}
\begin{aligned}
&egde\_score = (1-parallelism) * 0.8 + distance* 0.2 \\
&parallelism = |\frac{\mathbf{e_{static,i}}}{||\mathbf{e_{static,i}}||} \cdot \frac{\mathbf{e_{move,j}}}{||\mathbf{e_{move,j}}||}|,
\end{aligned}
\end{equation}
where $distance$ is calculated by sampling points on both edges and summing the distances between the points. 

However, when the two parts are in close contact, issues still arise. Assuming that the interaction location between the end-effector and the movable part is typically on the opposite side of the joint, we introduce the following criterion based on the above discussion:
\begin{equation}
\begin{aligned}
Edge\_Pair = \mathop{\arg\min}\limits_{i,j}(&\lambda_1egde\_score(\mathbf{e_{static,i}}, \mathbf{e_{move,j}})\\
&\pm \lambda_2 |\mathbf{e_{move,j}}\cdot \mathbf{v_{ee}}|\\
&\pm \lambda_3 distance(PC_{move}, \mathbf{e_{move,j}})),
\end{aligned}
\end{equation}
where $\pm$ corresponds to rotational joints and prismatic joints, respectively. Then the joint direction is calculated by:
\begin{equation}
joint\_direction = \frac{\mathbf{e_{move,Edge\_Pair[1]}}}{||\mathbf{e_{move,Edge\_Pair[1]}}||}
\end{equation}

We obtain the joint center through nearest points matching. In detail, for $P_1 \in Part_{move}$, find the closet point $P_2$ in $Part_{static}$ to establish a one-to-one mapping. We calculate the joint center by calculating the average of the midpoints of the $K$ closest point pairs near the $Edge_Part$:
\begin{equation}
joint\_center = \frac{1}{K}\sum_{k=1}^K(\frac{P_{1,k} + P_{2,k}}{2}),
\end{equation}
where:
\begin{equation}
\begin{aligned}
&[P_{1,k}, P_{2,k}] = Sort(||P_1 - P_2||, \text{~for}\\
&P_1 \in Part_{move} \& distance(P_1, \mathbf{e_{move,Edge\_Pair[1]}}))<\epsilon,\\
&P_2 \in Part_{static} \& distance(P_2, \mathbf{e_{static,Edge\_Pair[0]}}))<\epsilon.\\
\end{aligned}
\end{equation}

\subsubsection{Movable Part Motion Recovery}
\label{mr}
For articulated objects, the motion of the movable part can be described as the physical parameters. We use the trajectory of the contact points $Traj_r$ to supervise the movable part:
\begin{equation}
\label{eq: motion_replay}
\begin{aligned}
\min_{\theta_t} \, f(\theta_t) = \| PC_{r,t} - IP_t(Surf(Part_{move}(\theta_t)), Tacj_r) \|,
\end{aligned}
\end{equation}
where $IP(Surf(Part_{move}(\theta_t)), Tacj_r)$ calculates the intersection point between the plane formed by the surface of the movable part, which rotated by $\theta_t$ degrees or moved a distance of $\theta_t$, at time $t$ and the trajectory $Traj_r$.

\subsection{Articulated Object Replacement}
\label{ao_replace}
% To generate high-quality generalized data, the replacement of articulated objects must ensure: 1. The authenticity of appearance. 2. The accuracy of physical interactions. In this section, providing the same category of articulated object $AO_{new}$, we propose a novel process to automatically generating corresponding manipulation data, as illustrated in the right part of Fig~\ref{fig:pipeline}. Meanwhile, to expand the generalization boundary, the pose of the replacement object can be arbitrarily modified, and the corresponding robot arm joint states can be generated based on the trajectory interpolation.
Providing the same category of articulated object $AO_{new}$, we propose a novel process to process $AO_{new}$, including the geometry and appearance, as illustrated in the right part of Fig~\ref{fig:pipeline}. After processing the given articulated object asset, we convert the USD file into a 3DGS representation~\cite{scolari2025mesh2splat}. Then, the motion simulation of the new movable part can be obtained using the method from Section~\ref{mr}. Moreover, to expand the generalization boundary, the pose of the replacement object can be arbitrarily modified.

\subsubsection{Physical Interaction Adaptation}
% Given the USD $AO_{new}$, we first need to preprocess it to ensure its geometric accuracy when applied to the existing 3DGS scene $G^{'}$. To achieve it, we design a two-stage optimization approach using contact points for supervision to perform geometric processing on $AO_{new}$. 
Given the replacement digital asset for the articulated object $AO_{new}$, we design a two-stage optimization approach using contact points for supervision to perform geometric processing, ensuring the physical interaction between the new object and the original robotic arm trajectory is consistent and reasonable.

To get the accurate contact point of $AO_{new}$, a contact point mapping method inspired by NOCS~\cite{Wang_2019_CVPR} is proposed. Specifically, the point clouds $Part_{move}$ and $Part_{move}^{new}$ (obtained by sampling points from the mesh surface) are normalized to a unit cube space. Then, the relative position of $PC_{move}$ in the normalized space is projected onto the surface of the movable part of the newly normalized articulated object. Finally, by applying denormalization, the contact point information of the new articulated object $PC_{map}$ is obtained.

(1) For Stage 1, the following optimization function are used to obtain a rough estimate of the geometric features of the new articulated object:
\begin{equation}
\min_{\mathbf{g}} \, f_1(\mathbf{g}) = \sum_{t=start\_frame}^{end\_frame} \|PC_{r,t} - R_t(PC_{map} | \mathbf{g}) \|,
\end{equation}
where $\mathbf{g}=[s,r_{init},offset_{\{x,y\}}]$ represents the scaling, initial motion parameter of the movable part, and the XY-axis offset in the original object's coordinate system with respect to $AO_{new}$. Assuming the movable part moves at a constant speed, $R_t(PC_{map} | g)$ represents the contact point position of $PC_{map}$ after being transformed by $(r_{init} / (end\_frame-start\_frame))*t$ in the $AO_{new}$, following the application of the $g$ parameters. It is worth noting that $R_t(\cdot)$ corresponds to rotational motion for rotational joints and translational motion for prismatic joints.

(2) However, there are two issues in real-world operations that lead to optimization errors: 1. The movement of the movable part is not uniform. 2. There is some sliding between the end-effector contact point and the surface of the movable part, causing the trajectory of $PC_r$ denoted as $Traj_r$ to not be an arc. Therefore, we add a stage two optimization to minimize this error as much as possible. Like Equation~\ref{eq: motion_replay}, the specific optimization objective is as follows:
\begin{equation}
\begin{aligned}
\min_{\mathbf{g}} \, f_2(\mathbf{g}) &= \sum_{t=start\_frame}^{end\_frame} \| PC_{r,t} - \\
& \quad IP_t(Surf(Part_{move}(\mathbf{g})), Tacj_r) \|,
\end{aligned}
\end{equation}
% where $IP(Surf(Part_{move}(\mathbf{g})), Tacj_r)$ calculates the intersection point between the plane formed by the surface of the movable part at time $t$ and the trajectory $Traj_r$. Thus, we obtain the optimized geometric parameters $\mathbf{g}^{*}$ and apply them to the USD $AO_{new}$.
where $Part_{move}(\mathbf{g})$ stands for the point cloud of the movable part after the $\mathbf{g}$ transformation. Thus, we obtain the optimized geometric parameters $\mathbf{g}^{*}$ and apply them to the USD $AO_{new}$.

\subsubsection{Visual Enhancement}
At the same time, the visual performance of the replaced object is also crucial for the reality of the generated data. We apply the lighting changes from the real scene to the USD. First, we use DiffusionLight~\cite{Phongthawee2023DiffusionLight} to extract the lighting from the real environment. Then, bake the ambient lighting onto the object's material in Blender. Moreover, Gaussian Inpainting is applied to handle the Gaussian holes caused by object replacement.

\subsubsection{Pose Generalization}
Last, to further expand the generalization boundary, the pose of the articulated object can be arbitrarily transformed, denoted as $T_{ao}$. Begin with the existing trajectory of the end effector $\tau \in R^{7\times T}$, and divide it into three stages $\tau_{1,2,3}$ based on the $start\_frame$ and $end\_frame$. Since $\tau_2$ is relatively invariant with respect to the object, the new trajectory for the second stage is $\tau_2^{'}=T_{ao}\tau_2$. Given the trajectory $\tau_0 = {p_{start},\cdots,p_{end}}$, the new trajectory has the same starting pose $p_{start}$ but a new ending pose $p_{end}^{*}$. The translation of the new trajectory at different time frames is obtained using linear interpolation, while the rotation angles are obtained through spherical linear interpolation. After obtaining the new trajectory, the corresponding joint angle transformations can be obtained through inverse kinematics.
\section{Experiments}
\begin{figure}[t]  
    \centering  
    \begin{subfigure}[b]{0.45\textwidth}
    \centering
    \includegraphics[width=\textwidth]{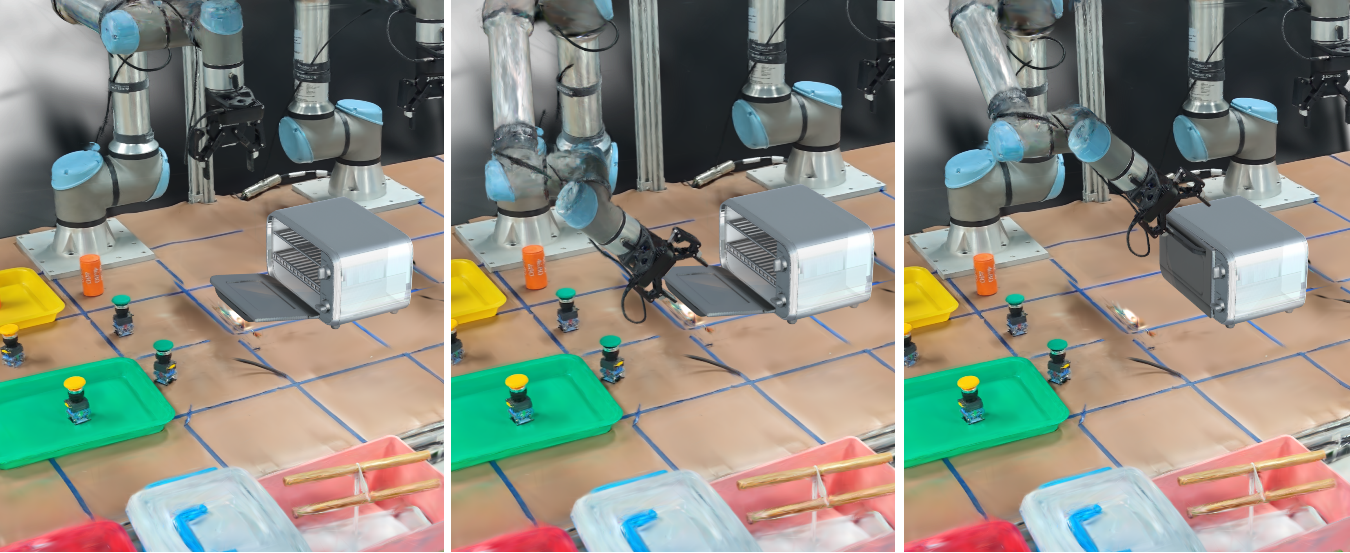}
    \caption{Microwave Oven}
    \label{fig:mic_data}
    \end{subfigure}
    \begin{subfigure}[b]{0.45\textwidth}
    \centering
    \includegraphics[width=\textwidth]{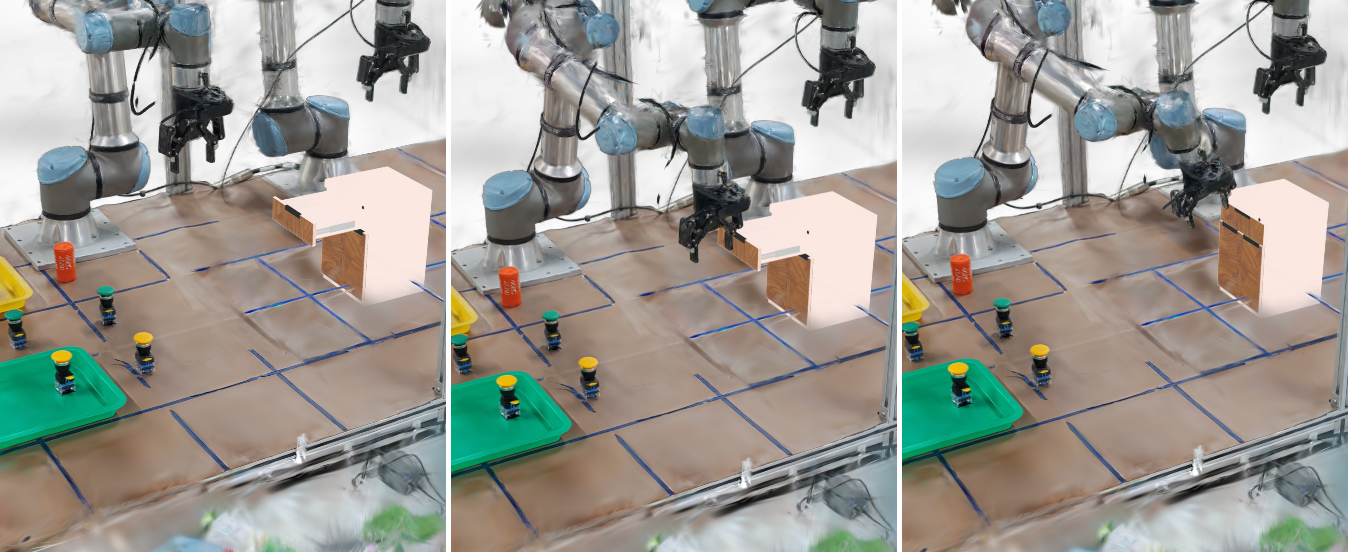}
    \caption{Drawer}
    \label{fig:drawer_data}
    \end{subfigure} 
    \caption{The Gaussian field visualizations of the data generated from AOMGen.} 
    \label{fig:data_vis} 
\end{figure}
We conduct a series of experiments to validate the effectiveness of the data generated by AOMGen, for rotational joints and prismatic joints. Figure~\ref{fig:data_vis} shows the Gaussian field visualizations after replacing the rotational and prismatic joint objects. Specifically, The experimental results address the following three questions:
\begin{itemize}
    \item Does the data generated by AOMGen align with real physical interactions?
    \item Can the manipulation data generated from AOMGen be directly used for VLA training?
    \item Can the generalization strategy enhance the robustness of the VLA model?
\end{itemize}
Next, detailed experimental settings and result analysis will be provided to answer the above questions.

\begin{table*}[t]
\centering
\begin{tabular}{l|ccccc|c}
\hline
\textbf{Replacement Object} & \textbf{Microwave Oven} & \textbf{Tool Box} & \textbf{Computer} & \textbf{Drawer} &\textbf{Cabinet} &\textbf{Average} \\ \hline
\textbf{Success Rate} & 98\% & 96\% & 96\% & 100\% & 100\% & 98\% \\ \hline
\end{tabular}
\caption{Success Rate in simulation replay for different articulated objects, including rotational joint and prismatic joint.}
\label{table:sr_replay}
\end{table*}

\subsection{Experiments Setup}
We collect the real demonstration data using a Universal Robot UR5e equipped with a 2F85 gripper. A mobile device is responsible for scanning the static scene, while a fixed camera records RGB images of the manipulation process.

To demonstrate the robustness of AOMGen in handling a variety of articulated objects, we select three different simulation assets with rotational joint, including microwave ovens, tool box and computer, and two assets with prismatic joint, including drawer and cabinet from ArtVIP~\cite{jin2025artviparticulateddigitalassets} for substitution. During pose generalization, the translation range of the object is $[-0.05m,0.3m] * [-0.05m,0.05m]$, and the rotation range is $[-45 ^{\circ}, 45 ^{\circ}]$. 

The entire model training is carried out on an NVIDIA RTX4090 GPU. Specially, the model $\pi_{0.5}$~\cite{intelligence2025pi05visionlanguageactionmodelopenworld} is fine-tuned using LoRA, with a batch size of 16 and a learning rate of $5\times10^{-5}$, employing cosine decay with a 10K step warm-up. The training consists of 30,000 steps, and the optimizer used is AdamW with gradient clipping set to 1.0. The action prediction horizon is set to 5 steps, and precision is configured with bfloat16 (frozen) and float32 (trainable). The model OpenVLA~\cite{kim24openvla} is fine-tuned by LoRA using 50,000 steps, enabled with a rank of 32. Moreover, we use IsaacSim as our simulation platform. 

\begin{figure}[b]  
    \centering  
    \begin{subfigure}[b]{0.45\textwidth}
    \centering
    \includegraphics[width=\textwidth]{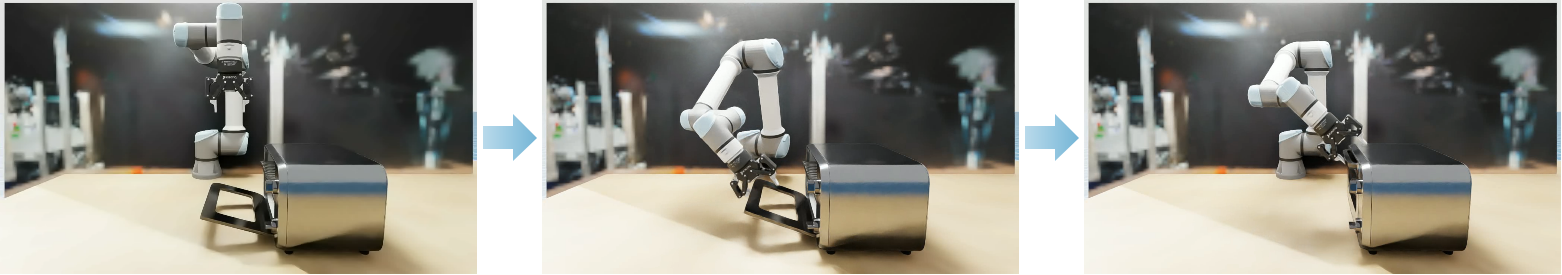}
    \caption{Microwave Oven}
    \label{fig:rotation_replay}
    \end{subfigure}
    \begin{subfigure}[b]{0.45\textwidth}
    \centering
    \includegraphics[width=\textwidth]{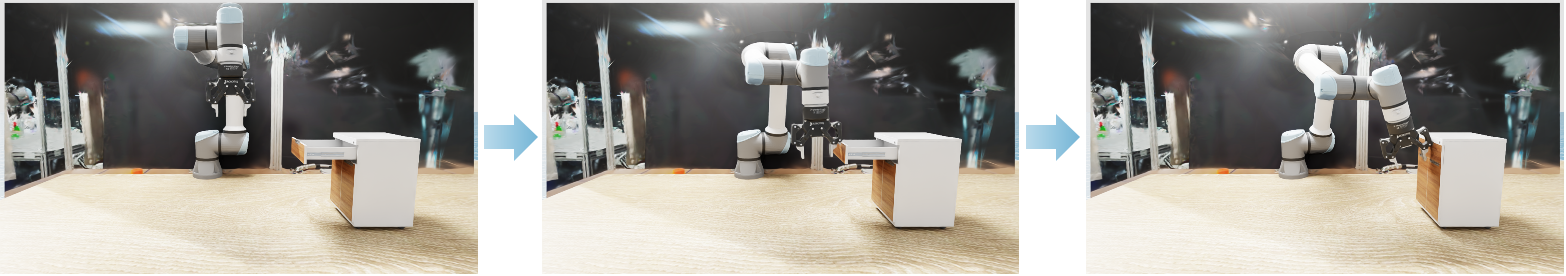}
    \caption{Drawer}
    \label{fig:prismatic_rep}
    \end{subfigure} 
    \caption{Replay of generated data in the simulator.} 
    \label{fig:replay} 
\end{figure}

\subsection{Simulator Replay}
To address Question 1, we replay the robotic arm trajectory and the target articulated object's pose generated from the data in the simulator to validate the physical interaction realism in the simulator. Specially, we import the optimized USD assets and their corresponding poses in Sec~\ref{ao_replace} into the simulator, while replaying the robot arm's joint states at different time steps, and observe whether the corresponding tasks are completed. 50 data for each replacement object entries with only pose changes are generated, and their Success Rate (SR) in simulation is calculated in Table~\ref{table:sr_replay}. The results of the simulator replay demonstrate that the generated data adheres to basic physical laws, providing a fundamental guarantee for the effectiveness of subsequent model training. AOMGen can generate physically plausible demonstration data with various replacement objects of the same category. The Figure~\ref{fig:replay} shows the replay process of a data sample.

\subsection{VLA Training}
\label{sec::vla_train}
In order to answer Question 2, we compare three policies that are respectively trained on demonstrations generated for each replacement object:
\begin{enumerate}
\item The original model, without any data fine-tuning.
\item Fine-tuning is performed using 50 data samples generated by AOMGen.
\item Fine-tuning is performed using 150 data samples generated by AOMGen.
\end{enumerate}
For each replacement object, during the inference phase, we import the simulation assets of the robotic arm, the articulated object, and the table. To reduce the gap between the simulation environment and the real world, we add a 'background wall' in the simulation environment and apply 3DGS render images, which have already undergone coordinate frame alignment, as textures on the background wall to simulate the real operating scene. To ensure the validity of the experimental results, we conduct experiments on two different models: $\pi_{0.5}$~\cite{intelligence2025pi05visionlanguageactionmodelopenworld} and OpenVLA~\cite{kim24openvla}. What's more, each evaluation is conducted with 30 trials.

\begin{figure*}[htbp]  
    \centering  
    \includegraphics[width=0.9\textwidth]{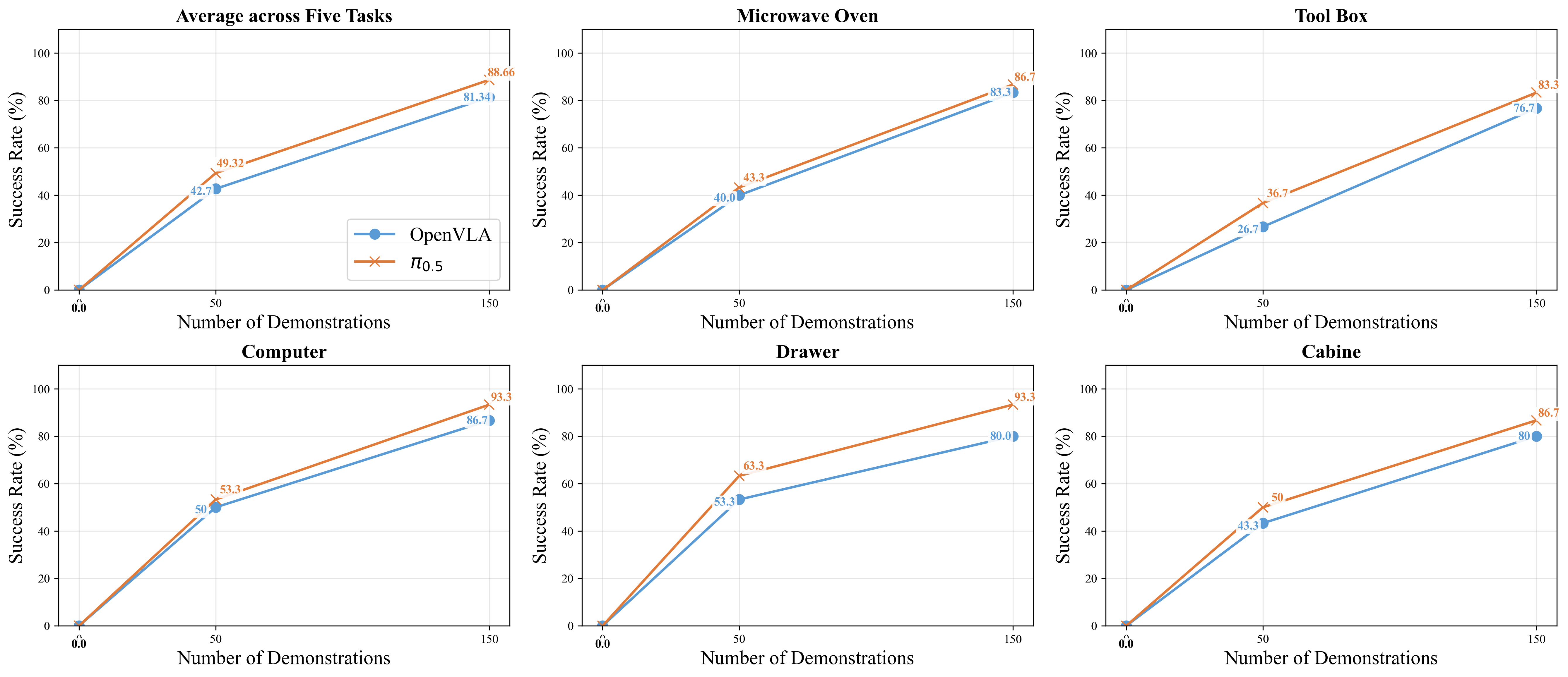} 
    \caption{Success rate of the fine-tuned VLA model under different object replacements.} 
    \label{fig:vla_sr} 
\end{figure*}

As shown in Fig~\ref{fig:vla_sr}, for each model, as the number of generated demonstrations increases, a remarkable improvement in success rate is observed. Specially, the performance of $\pi_{0.5}$ achieves 88.66, and the performance of OpenVLA reaches 81.34, which reflects that after fine-tuning with AOMGen-generated articulated object manipulation data, the model has acquired the ability to manipulate the corresponding object.

\subsection{Robustness Analysis}
\begin{figure*}[htbp]
  \centering
  
  % 第一个子图
   \begin{subfigure}[b]{0.3\textwidth}
    \centering
    \includegraphics[width=\textwidth]{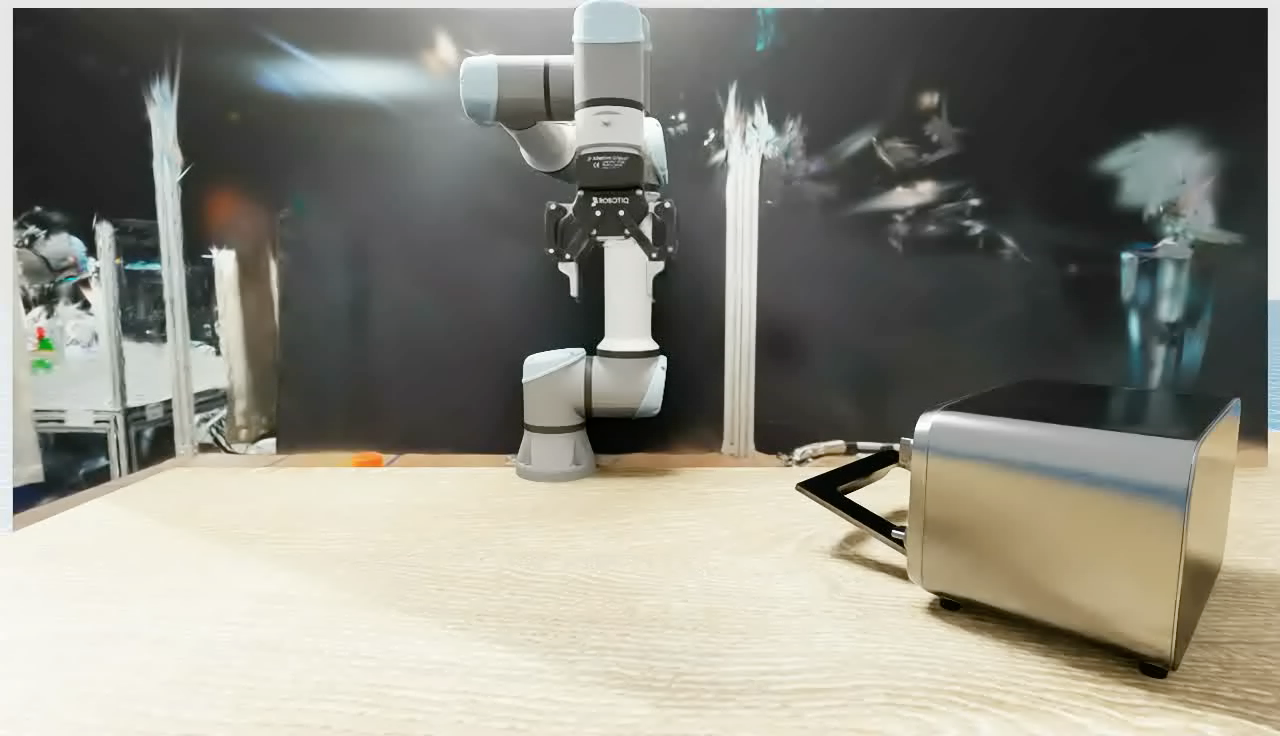}
    \caption{Pose Generalization}
    \label{fig:sub2}
  \end{subfigure}
  \hfill
  % 第二个子图
  \begin{subfigure}[b]{0.3\textwidth}
    \centering
    \includegraphics[width=\textwidth]{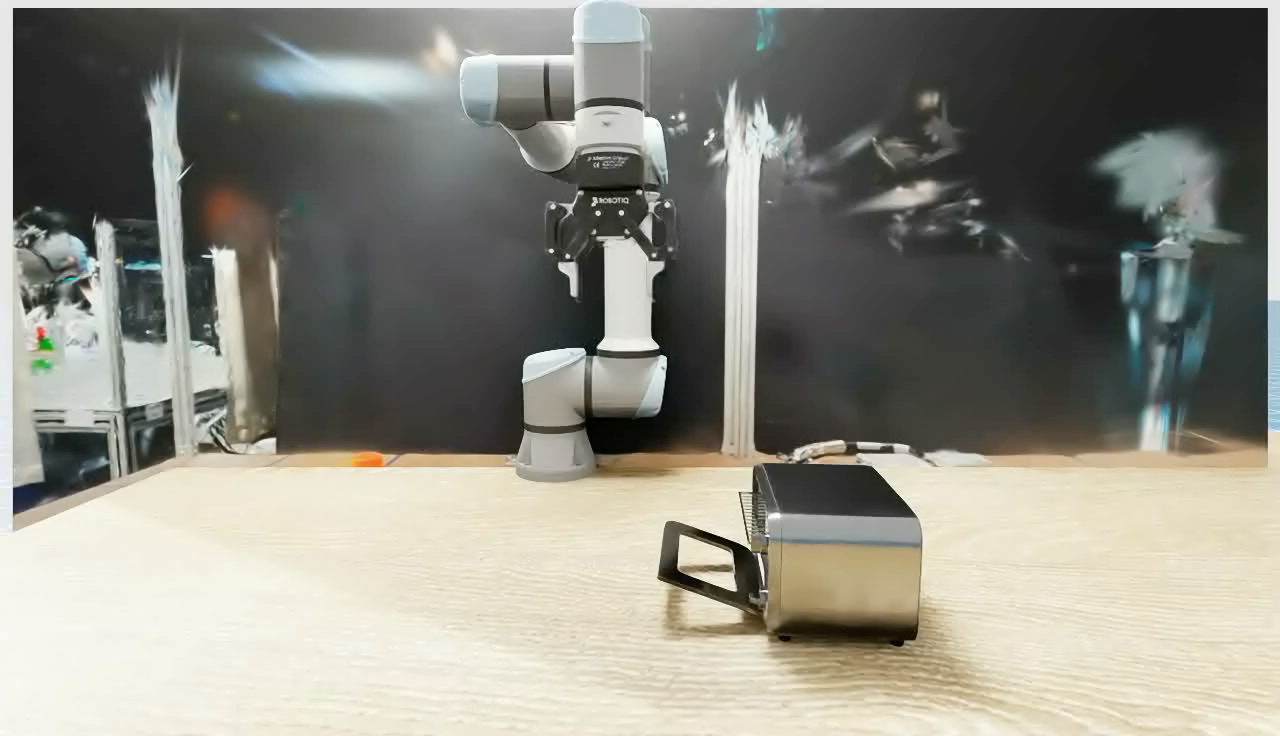}
    \caption{Scale Generalization}
    \label{fig:sub1}
  \end{subfigure}
  \hfill
  % 第三个子图
  \begin{subfigure}[b]{0.3\textwidth}
    \centering
    \includegraphics[width=\textwidth]{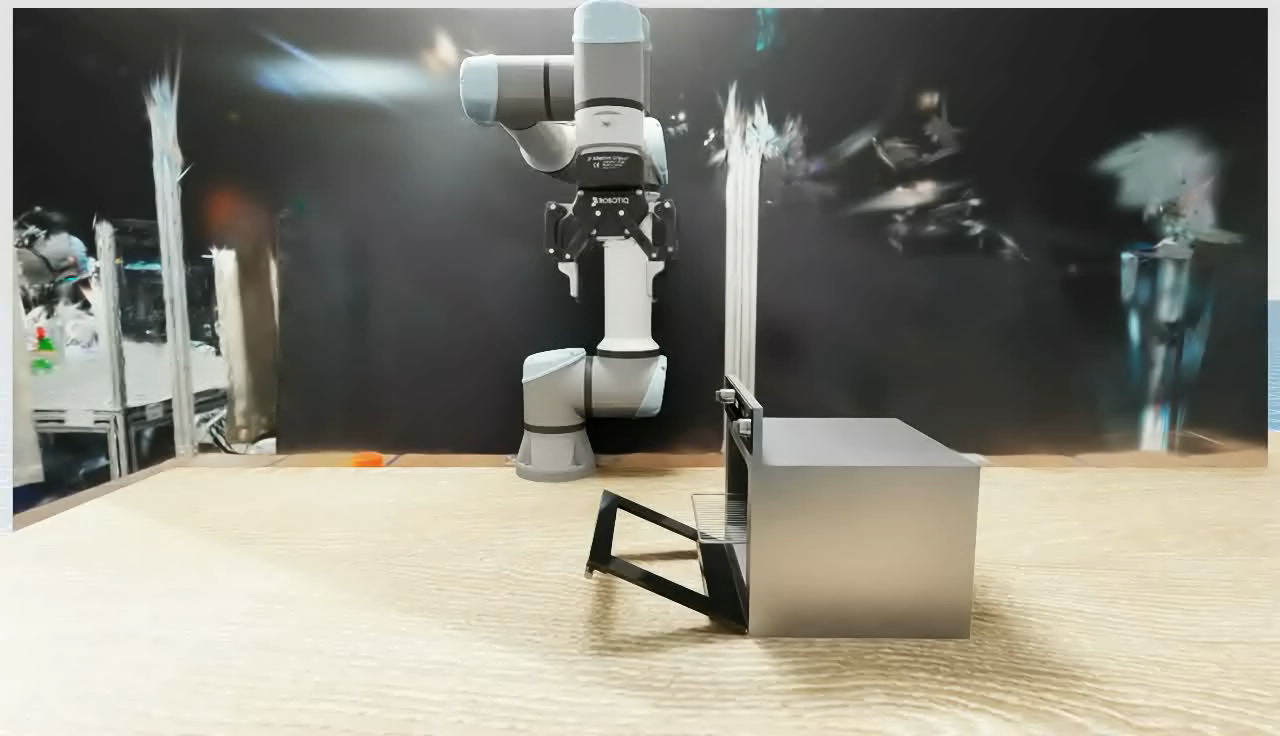}
    \caption{Articulated Object Genelization}
    \label{fig:sub3}
  \end{subfigure}
  
  \caption{Demonstrations of robustness experiments for different articulated objects.}
  \label{fig:overall}
\end{figure*}

In fact, due to the differences between 3DGS rendered images and the simulator environment, the experimental results in Sec. 2 already reflect, to some extent, the model's robustness to the environment. At the same time, the models are tested under different initial object position configurations, demonstrating their robustness to pose variations. Moreover, it validates the fundamental differences between the data generated by AOMGen and the data generated by the simulator. In this section, we describe how the powerful generalization capability of AOMGen is reflected in the model's robustness. Illustrative examples of the VLA robustness experiments in various aspects are shown in Fig~\ref{fig:overall}.

For rotational joints and translational joints, we follow the fine-tuning method described in Sec~\ref{sec::vla_train} to obtain the models of OpenVLA and $\pi_{0.5}$ fine-tuned with 150 data samples, with different sizes of the object. First, we perform scale generalization on the optimized USD to test the fine-tuned model's robustness. We select the microwave oven and drawer for the experiment, and its optimized USD asset is progressively scaled between 0.6 and 0.9 to test the VLA model fine-tuned with data generated by AOMGen in the simulator. Each configuration is repeated 20 times to test the success rate. The results are listed in the Table~\ref{table:scale_robust}. As the results shown in the table, simple processing of the data enables the model to handle the same object at different scales.

\begin{table}[htbp]
\footnotesize
  \centering
  \setlength{\tabcolsep}{4.0 pt}
  \begin{tabular}{c|c|cccc}  % 保留 5 列，去掉最后一列
    \toprule
    \multirow{3}{*}{\textbf{Object}} & \multirow{3}{*}{\textbf{VLA Model}} & \multicolumn{4}{c}{\textbf{Success Rate}}\\ 
     &   & 0.6  &  0.7 & 0.8 & 0.9 \\  \midrule
     \multirow{2}{*}{\textbf{Microwave Oven}} & {OpenVLA}  & 55 & 65 & 65 & 70 \\
      & {$\pi_{0.5}$}  & 65 & 75 & 80 & 85 \\
      \midrule
     \multirow{2}{*}{\textbf{Drawer}} & {OpenVLA}  & 60 & 70 & 80 & 80 \\
      & {$\pi_{0.5}$} & 80 & 90 & 90 & 90 \\
      \bottomrule
    \end{tabular}
  \caption{Performance when changing the scale of given replacement articulated object.}
    \label{table:scale_robust}
     \vspace{-3mm}
\end{table}

Then, we validate whether the model is capable of handling with unseen objects after training on mixed data with different replacement articulated objects. Specifically, for rotational joint, we fine-tune the VLA model using both single data and mixed data. Then, an unseen object is selected. 20 trails are conducted to test the task success rate. The results are shown in the Table~\ref{table:unseen_object}. We find that the VLA model fine-tuned with single data still lacks generalization ability on unseen objects. However, the mixed data generated through object replacement helps the model successfully handle unseen objects of the same category. This demonstrates the advantage of AOMGen in arbitrarily replacing objects of the same category and generating corresponding data for downstream tasks.

\begin{table}[htbp]
\centering
\begin{tabular}{lc}
\hline
\textbf{Train Data } & Unseen Object \\ \hline
\textbf{Single Data} &  15\% \\ 
\textbf{Mixed Data} &  65\% \\ 
\hline
\end{tabular}
\caption{Success Rate with model trained by different dataset, to verify the model's task success rate on unseen objects.}
\label{table:unseen_object}
\end{table}
\section{Conclusion}
In this paper, we propose AOMGen, a powerful articulated object manipulation data generator that includes both rotational joint and prismatic joint. A series of novel designs enhance the physical interaction accuracy of the generated data and the visual realism of the rendered images. The generated data is verified to be applicable for fine-tuning the VLA model, further enhancing the model's robustness.
{
    \small
    \bibliographystyle{ieeenat_fullname}
    \bibliography{main}
}


\end{document}